\documentclass[10pt,twocolumn,letterpaper]{article}

\usepackage{cvpr}
\usepackage{times}
\usepackage{epsfig}
\usepackage{graphicx}
\usepackage{amsmath}
\usepackage{amssymb}
\usepackage[]{algorithm2e}

\usepackage[pagebackref=true,breaklinks=true,letterpaper=true,colorlinks,bookmarks=false]{hyperref}

\usepackage{url}
\usepackage{graphicx}
\graphicspath{{imgs/}}
\DeclareGraphicsExtensions{.pdf,.jpg,.png}
\usepackage{xcolor}

\usepackage{arydshln}

\usepackage{enumitem}
\setlist{nolistsep}

\newcommand{\fakesubsection}[1]{\smallskip\noindent\textbf{#1:}}

\cvprfinalcopy 

\def\cvprPaperID{462} 

\ifcvprfinal\fi
\begin{document}

\title{Temporal Convolutional Networks: \\A Unified Approach to Action Segmentation}

\author{Colin Lea 
	\hspace{11px} Ren\'{e} Vidal  
	\hspace{11px} Austin Reiter
	\hspace{11px} Gregory D. Hager   \\	
Johns Hopkins University \\
	 \{clea1@, rvidal@cis., areiter@cs., hager@cs.\}jhu.edu
}


\maketitle
\begin{abstract}

The dominant paradigm for video-based action segmentation is composed of two steps:
first, for each frame, compute low-level features using Dense Trajectories or a Convolutional Neural Network that encode spatiotemporal information locally, and second, input these features into a classifier that captures high-level temporal relationships, such as a Recurrent Neural Network (RNN). 
While often effective, this decoupling requires specifying two separate models, each with their own complexities, and prevents capturing more nuanced long-range spatiotemporal relationships.
We propose a unified approach, as demonstrated by our Temporal Convolutional Network (TCN), that hierarchically captures relationships at low-, intermediate-, and high-level time-scales.
Our model achieves superior or competitive performance using video or sensor data on three public action segmentation datasets and can be trained in a fraction of the time it takes to train an RNN.

\end{abstract}
\section{Introduction}
\label{sec:intro}

Action segmentation is crucial for numerous applications ranging from collaborative robotics to modeling activities of daily living. 
Given a video, the goal is to simultaneously segment every action in time and classify each constituent segment.
While recent work has shown strong improvements on this task, 
models tend to decouple low-level feature representations from high-level temporal models. 
Within video analysis, these low-level features may be computed by pooling handcrafted features (e.g. Improved Dense Trajectories (IDT)~\cite{wang_iccv_2013}) 
or concatenating learned features (e.g. Spatiotemporal Convolutional Neural Networks (ST-CNN)  \cite{lea_eccv_2016,ng_cvpr_2015}) over a short period of time. 
High-level temporal classifiers capture a local history of these low-level features. In a Conditional Random Field (CRF), the action prediction at one time step is are often a function of the prediction at the previous time step, and in a Recurrent Neural Network (RNN), the predictions are a function of a set of latent states at each time step, where the latent states are connected across time. 
This two-step paradigm has been around for decades (e.g., \cite{hofmann_accel_1997}) and typically goes unquestioned. However, we posit that valuable information is lost between steps. 


\begin{figure}
	\center
	\includegraphics[width=\hsize]{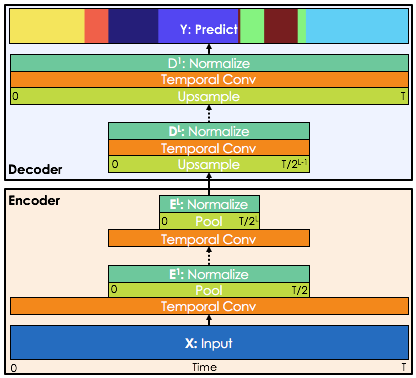}
	\caption{Our temporal encoder-decoder network hierarchically models actions from video or other time-series data.}
	\label{fig:unified_model}
\end{figure}

In this work, we introduce a unified approach to action segmentation that uses a single set of computational mechanisms -- 1D convolutions, pooling, and channel-wise normalization -- to hierarchically capture low-, intermediate-, and high-level temporal information. 
For each layer, 1D convolutions capture how features at lower levels change over time, pooling enables efficient computation of long-range temporal patterns, and normalization improves robustness towards various environmental conditions.
In contrast with RNN-based models, which compute a set of latent activations that are updated sequentially per-frame, we compute a set of latent activations that are updated hierarchically per-layer. 
As a byproduct, our model takes much less time to train.
Our model can be viewed as a generalization of the recent ST-CNN~\cite{lea_eccv_2016}
and is more similar to recent models for semantic segmentation than it is to models for video-analysis.
We show this approach is broadly applicable to video and other types of robot sensors.

\fakesubsection{Prior Work}
Due to space limitations, here we will only briefly describe models for time-series and semantic segmentation. See~\cite{lea_eccv_2016} for related work on action segmentation or~\cite{vrigkas_frontiers_2015} for a broader overview on action recognition. 


RNNs and CRFs are popular high-level temporal classifiers. 
RNN variations, including Long Short Term Memory (LSTM) and Gated Recurrent Units (GRU), model hidden temporal states via internal gating mechanisms. However, they are hard to introspect and difficult to correctly train~\cite{pascanu_icml_2013}.
It has been shown that in practice LSTM only keeps a memory of about 4 seconds on some video-based action segmentation datasets~\cite{singh_cvpr_2016}.
CRFs typically model pairwise transitions between the labels or latent states (e.g.,~\cite{lea_eccv_2016}), 
which are easy to interpret, but over-simplify the temporal dynamics of complex actions.
Both of these models suffer from the same fundamental issue: intermediate activations are typically a function of the low-level features at the current time step and the state at the previous time step. Our temporal convolutional filters are a function of raw data across a much longer period of time.

Until recently, the dominant paradigm for semantic semantic was similar to that of action segmentation. Approaches typically
combined low-level texture features (e.g., TextonBoost) with high-level spatial models (e.g., grid-based CRFs) that model the relationships between different regions of an image~\cite{krahenbuhl_nips_2011}. 
This is similar to action segmentation where low-level spatiotemporal features are used in tandem with high-level temporal models. 
Recently, with the introduction of Fully Convolutional Networks (FCNs), the dominant semantic segmentation paradigm has started to change.
Long \etal~\cite{long_cvpr_2015} introduced the first FCN, which leverages typical classification CNNs like AlexNet, to compute per-pixel object labels. This is done by intelligently upsampling the intermediate activations in each region of an image.
Our model is more similar to the recent encoder-decoder network by Badrinarayanan \etal~\cite{badrinarayanan_arxiv_2015}. Their encoder step uses the first half of a VGG-like network to capture patterns in different regions of an image and their decoder step takes the activations from the encoder, which are of a reduced image resolution, and uses convolutional filters to upsample back to the original image size. In subsequent sections we describe our temporal variation in detail. 
\section{Temporal Convolutional Networks (TCN)}
\label{sec:TCN}

The input to our Temporal Convolutional Network can be a sensor signal (e.g. accelerometers) or latent encoding of a spatial CNN applied to each frame. Let $X_t \in \mathbb{R}^{F_0}$ be the input feature vector of length $F_0$ for time step $t$ for $0 < t \leq T$. Note that the time $T$ may vary for each sequence, and we denote the number of time steps in each layer as $T_l$. The true action label for each frame is given by $y_t \in \{1,\dots,C\}$, where $C$ is the number of classes.

Our encoder-decoder framework, as depicted in Figure~\ref{fig:unified_model}, is composed of temporal convolutions, 1D pooling/upsampling, and channel-wise normalization layers.


For each of the $L$ convolutional layers in the encoder, we apply a set of 1D  filters that capture how the input signals evolve over the course of an action.
The filters for each layer are parameterized by tensor $W^{(l)} \in \mathbb{R}^{F_{l} \times d \times F_{l-1}}$ and biases $b^{(l)} \in \mathbb{R}^{F_{l}}$, where $l \in \{1,\dots,L\}$ is the layer index and $d$ is the filter duration. 
For the $l$-th layer of the encoder, the $i$-th component of the (unnormalized) activation $\hat{E}^{(l)}_t \in \mathbb{R}^{F_{l}}$ is a function of the incoming (normalized) activation matrix $E^{(l-1)} \in \mathbb{R}^{ F_{l-1} \times T_{l-1}}$ from the previous layer
\begin{align}
\hat{E}^{(l)}_{i,t} = f( b_i^{(l)} + \sum_{t'=1}^{d} \langle  W_{i,t',\cdot}^{(l)},  E^{(l-1)}_{\cdot,t+d-t'} \rangle )
\end{align}
for each time $t$ where $f(\cdot)$ is a Leaky Rectified Linear Unit. The normalization process is described below.

Max pooling  is applied with width 2 across time (in 1D) such that $T_l = \frac{1}{2} T_{l-1}$.\footnote{In theory, this implies $T$ must divisible by $2^L$. In practice, we pad each sequence to be of an appropriate length, given the pooling operations, such that the input length of the whole sequence, $T$, and the length of the output predictions are the same.}
Pooling enables us to efficiently compute activations over a long period of time.

We apply channel-wise normalization after each pooling step in the encoder. This has been effective in recent CNN methods including Trajectory-Pooled Deep-Convolutional Descriptors (TDD)~\cite{TDD}. We normalize the pooled activation vector $\hat{E}^{(l)}_{t}$ by the highest response at that time step, $m = \max_i \hat{E}^{(l)}_{i,t}$, with some small $\epsilon=1\text{\sc{e}-}5$ such that
\begin{align}
E^{(l)}_{t} = \frac{1}{m+ \epsilon} \hat{E}^{(l)}_{t}.
\end{align}

Our decoder is similar to the encoder, except that upsampling is used instead of pooling, and the order of the operations is now upsample, convolve, then normalize.  Upsampling is performed by simply repeating each entry twice. 



The probability that frame $t$ corresponds to one of the C action classes is predicted by vector $\hat Y_t \in [0,1]^C$ using weight matrix $U \in \mathbb{R}^{C \times F_0}$ and bias $c \in \mathbb{R}^{C}$
\begin{align}
\hat{Y}_t = \text{softmax}(U D^{(1)}_t + c).
\end{align}

We explored many other mechanisms, such as adding skip connections between layers, using different patterns of convolutional layers, and other normalization schemes. These helped at times and hurt in others. The aforementioned solution was superior in aggregate. 



\fakesubsection{Implementation details}
Each of the $L=3$ layers has $F_l=\{32,64,96\}$ filters. Filter duration, $d$, is set as the mean segment duration for the shortest class from the training set. For example, $d=10$ seconds for 50 Salads. 
Parameters of our model were learned using the cross entropy loss with Stochastic Gradient Descent and ADAM step updates. 
All models were implemented using Keras and TensorFlow.

For each frame in our video experiments, the input, $X_t$, is the first fully connected layer computed in a spatial CNN trained solely on each dataset. We trained the model of~\cite{lea_eccv_2016}, except instead of using Motion History Images (MHI) as input to the CNN, we concatenate the following for image $I_t$ at frame $t$: $[I_t, I_{t-d}-I_t, I_{t+d}-I_t, I_{t-2d}-I_t, I_{t+2d}-I_t]$ for $d=0.5$ seconds. 
In our experiments, these difference images -- which are a simple type of attention mechanism -- tend to perform better than MHI or optical flow across these datasets. 
Furthermore, for each time step, we perform channel-wise normalization before feeding it into the TCN. This helps with large environmental fluctuations, such as changes in lighting. 
\section{Evaluation}
\label{sec:evaluation}

We evaluate on three public datasets that contain action segmentation labels, video, and in two cases sensor data.




\textbf{University of Dundee 50 Salads ~\cite{stein_ubicomp_2013}}
 contains 50 sequences of users making a salad. Each video is 5-10 minutes in duration and contains around 30 action instances such as \texttt{cutting a tomato} or \texttt{peeling a cucumber}. This dataset includes video and synchronized accelerometers attached to ten objects in the scene, such as the \textit{bowl}, \textit{knife}, and \textit{plate}. We performed cross validation with 5 splits on the ``eval'' action granularity which includes 10 action classes.
Our sensor results used the features from \cite{lea_icra_2016} which are the absolute values of accelerometer values. 
Previous results (e.g., \cite{lea_icra_2016,richard_cvpr_2016}) were evaluated using different setups. For example, \cite{lea_icra_2016} smoothed out short interstitial background segments. We reran all results to be consistent with~\cite{richard_cvpr_2016}. We also included an LSTM baseline for comparison which uses $64$ hidden states.


\textbf{JHU-ISI Gesture and Skill Assessment Working Set (JIGSAWS) ~\cite{JIGSAWS}}
was introduced to improve quantitative evaluation of robotic surgery training tasks. 
We used Leave One User Out cross validation on the suturing activity, which consists of 39 sequences performed by 8 users about 5 times each. The dataset includes video and synchronized robot kinematics (position, velocity, and gripper angle) for each robot end effector as well as corresponding action labels with 10 action classes. Sequences are a few minutes long and typically contain around 20 action instances.

\textbf{Georgia Tech Egocentric Activities (GTEA)~\cite{fathi_cvpr_2011}}
contains 28 videos of 7 kitchen activities including making a sandwich and making coffee. For each of the four subjects, there is one instance of each activity.
The camera is mounted on the head of the user and is pointing at the area in front them. 
On average there are about 30 actions per video and videos are around a minute long. We used the 11 action classes defined in~\cite{fathi_iccv_2011} and evaluated using leave one user out. We show results for user 2 to be consistent with \cite{fathi_iccv_2011} and \cite{singh_cvpr_2016b}.



\fakesubsection{Metrics}
We evaluated using accuracy, which is simply the percent of correctly labeled frames, and segmental edit distance~\cite{lea_icra_2016}, which measures the correctness of the predicted temporal ordering of actions.
This edit score is computed by applying the Levenstein distance to the segmented predictions (e.g. $AAABBA \rightarrow ABA$). This is normalized to be in the range $0$ to $100$ such that higher is better. 

\begin{table*}[t]
	\centering
	\begin{tabular}{ccc}
		\begin{tabular}{| l | c | c |}
			\multicolumn{3}{c}{50 Salads (``eval'' setup)}\\
			\hline
			\textbf{Sensor-based}  & \textbf{Edit} & \textbf{Acc}\\
			\hline 		
			\cite{lea_icra_2016} LC-SC-CRF	 & 50.2 & 77.8 		\\	
			LSTM & 54.5 & 73.3\\ 			
			TCN   & \textbf{65.6} & \textbf{82.0}\\			
			\hline
			\textbf{Video-based} & \textbf{Edit} & \textbf{Acc} \\
			\hline
			\cite{lea_eccv_2016} VGG  & 7.6
			& 38.3 \\			
			\cite{lea_eccv_2016} IDT  & 16.8 & 54.3
			\\
			\cite{lea_eccv_2016}  Seg-ST-CNN  & \textbf{62.0} & 72.0\\ 
			Spatial CNN & 28.4  & 68.6 \\
			ST-CNN & 55.5 & 74.2 \\								
			TCN  & 61.1 & \textbf{74.4}\\
			\hline
		\end{tabular}&
		\begin{tabular}{| l | c | c |}
			\multicolumn{3}{c}{GTEA}\\
			\hline
			\textbf{Video-based} & \textbf{Edit} & \textbf{Acc} \\
			\hline
			\cite{fathi_iccv_2011} Hand-crafted & - & 47.7  \\			
			\cite{singh_cvpr_2016b} EgoNet & - & 57.6  \\				
			\cite{singh_cvpr_2016b} TDD & - & 59.5 \\
			\cite{singh_cvpr_2016b} EgoNet+TDD & - & \textbf{68.5} \\			
			Spatial CNN  & 36.6 & 56.1  \\
			ST-CNN  & 53.4 & 64.5 \\			
			TCN & \textbf{58.8} & 66.1  \\
			\hline
		\end{tabular}		
		&
		\begin{tabular}{| l | c | c |}
			\multicolumn{3}{c}{JIGSAWS}\\
			\hline
			\textbf{Sensor-based}  & \textbf{Edit} & \textbf{Acc} \\
			\hline 
			\cite{dipietro_miccai_2016} LSTM  & 75.3 & 80.5\\ 			
			\cite{lea_icra_2016} LC-SC-CRF & 76.8 & \textbf{83.4}  \\						
			\cite{dipietro_miccai_2016} Bidir LSTM  & 81.1 & 83.3\\ 

			\cite{Stefati_m2cai_2015} SD-SDL & 83.3  & 78.6 \\
			TCN  & \textbf{85.8} & 79.6\\			
			
			\hline 
		
			\textbf{Vision-based }  & \textbf{Edit} & \textbf{Acc}\\
			\hline
			\cite{tao_miccai_2013} MsM-CRF  & -& 71.7 \\			
			\cite{lea_eccv_2016} IDT  & 8.5 & 53.9 \\
			\cite{lea_eccv_2016} VGG  & 24.3 & 45.9  \\ 				
			\cite{lea_eccv_2016} Seg-ST-CNN &  66.6 & 74.7\\ 
			Spatial CNN & 37.7 & 74.0\\
			ST-CNN & 68.0 & 77.7\\
			TCN  & \textbf{83.1} & \textbf{81.4}\\			
			\hline
		\end{tabular}
	\end{tabular}
	\caption{ Results on 50 Salads, Georgia Tech Egocentric Activities, and  JHU-ISI Gesture and Skill Assessment Working Set. Notes: (1) Results using VGG and Improved Dense Trajectories (IDT) were intentionally computed without a temporal component for ablative analysis, hence their low edit scores. (2) We re-computed~\cite{lea_icra_2016} using the author's public code to be consistent with the setup of~\cite{richard_cvpr_2016}.}
	\label{table:results}
\end{table*}

\section{Experiments and Discussion}
Table~\ref{table:results} includes results for all datasets and corresponding sensing modalities. We include results from the spatial CNN which is input into the TCN, the Spatiotemporal CNN of Lea \etal~\cite{lea_eccv_2016} applied to the spatial features, and our TCN.

One of the most interesting findings is that some layers of convolutional filters appear to learn temporal shifts. There are certain actions in each dataset which are not easy to distinguish given the sensor data. 
By visualizing the activations for each layer, we found our model surmounts this issue by learning temporal offsets from activations in the previous layer. 
In addition, we find that despite the fact that we do not use a traditional temporal model, such as an RNN or CRF, our predictions do not suffer as heavily from issues like over-segmentation. This is highlighted by the large increase in edit score on most experiments. 

Richard \etal~\cite{richard_cvpr_2016} evaluated their model on the mid-level action granularity of 50 Salads which has 17 action classes. Their model achieved 54.2\% accuracy, 44.8\% edit, 0.379 mAP IoU overlap with a threshold of 0.1, and 0.229 mAP with a threshold of 0.5.\footnote{We computed our metrics using the predictions given by the authors.}
Our model achieves 59.7\% accuracy, 47.3\% edit, 0.579 mAP at 0.1, and 0.378 mAP at 0.5.

On GTEA, Singh \etal~\cite{singh_cvpr_2016b} reported 64.4\% accuracy by performing cross validation on users 1 through 3. We achieve 62.5\% using this setup. We found performance of our model has high variance between different trials on GTEA  -- even with the same hyper parameters -- thus, the difference in accuracy is not likely to be statistically significant. Our approach could be used in tandem with features from Singh \etal to achieve superior performance.





Our model can be trained much faster than an RNN-LSTM. 
Using an Nvidia Titan X, it takes on the order of a minute to train a TCN for each split, whereas it takes on the order of an hour to train an RNN-LSTM.
The speedup comes from the fact that we compute one set of convolutions for each layer, whereas RNN-LSTM effectively computes one set of convolutions for each time step. 



\fakesubsection{Conclusion}
We introduced a model for action segmentation that learns a hierarchy of intermediate feature representations, which contrasts with the traditional low- versus high-level paradigm. This model achieves competitive or superior performance on several datasets and can be trained much more quickly than other models. 
A future version of this manuscript will include more comparisons and insights on the TCN.

{\small
\bibliographystyle{ieee}
\bibliography{bib/abrevs_short,bib/ColinLea,bib/activity_recognition,bib/surgical}
}

\end{document}